\documentclass[letterpaper, 10 pt, conference]{ieeeconf}  

\IEEEoverridecommandlockouts                              

\usepackage[english]{babel}
\usepackage{amsmath}
\usepackage{amssymb}
\usepackage{booktabs}
\usepackage{caption}
\usepackage{subfigure}
\usepackage[dvips]{graphicx}
\usepackage{multirow}
\usepackage{amsfonts}
\usepackage{breakurl}
\usepackage{enumerate}
\usepackage{tabularx}
\usepackage{algorithm,algorithmic}
\usepackage{bm}
\usepackage{enumerate}
\usepackage{adjustbox}
\usepackage{xcolor}
\usepackage{siunitx}
\usepackage{booktabs}
\usepackage{mdframed}
\usepackage{adjustbox}
\usepackage{authblk}
\usepackage{color}
\usepackage{xurl}
\usepackage{cite}
\makeatletter
\let\NAT@parse\undefined
\makeatother
\usepackage{hyperref}
\usepackage{relsize}
\usepackage{float}
\usepackage{pifont}

\usepackage{stackengine,scalerel}

\usepackage{fontawesome5}

\title{\LARGE \bf Unifying Foundation Models with Quadrotor Control for\\Visual Tracking Beyond Object Categories}

\author{Alessandro Saviolo$^{*}$, Pratyaksh Rao$^{*}$, Vivek Radhakrishnan, Jiuhong Xiao, and Giuseppe Loianno
\thanks{$^*$These authors contributed equally.}
\thanks{The authors are with the New York University, Tandon School of Engineering, Brooklyn, NY 11201, USA. {\tt\footnotesize email: \{as16054, pr2257, vr2171, jx1190, loiannog\}@nyu.edu}.}
\thanks{This work was supported by the NSF CAREER Award 2145277, the DARPA YFA Grant D22AP00156-00, Qualcomm Research, Nokia, and NYU Wireless.}
}

\begin{document}

\thispagestyle{empty}
\pagestyle{empty}
\makeatletter
\g@addto@macro\@maketitle{
    \setcounter{figure}{0}
    \centering
    \includegraphics[width=\linewidth, trim=10 370 10 0, clip]
    {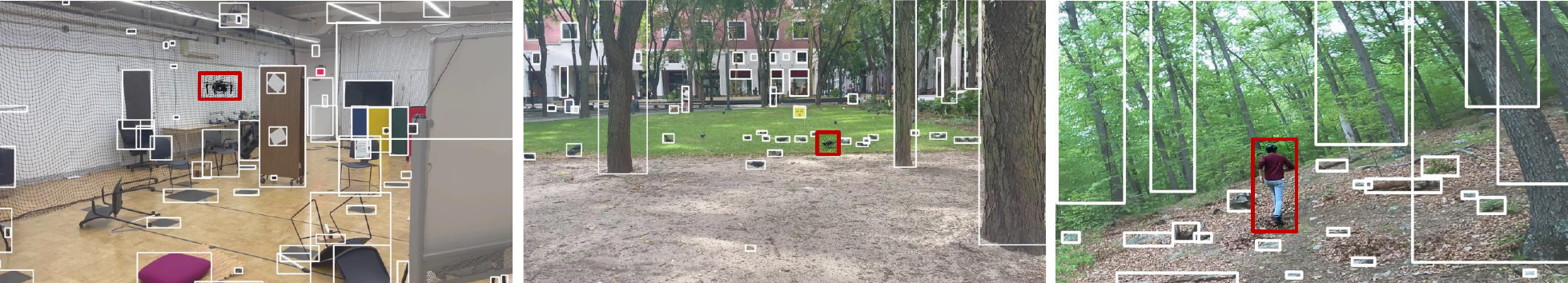}
    \captionof{figure}{
    From cluttered indoor to urban and rural outdoor settings, our foundation detector precisely detects a range of objects from humans to pigeons and custom drones  (white). The tracker, initially prompted by the user, maintains visibility of the desired object (red), while the visual controller swiftly navigates the quadrotor toward the target.
    \label{fig:initial_figure}
        \vspace{-10pt}
}
}
\makeatother
\maketitle


\begin{abstract}
Visual control enables quadrotors to adaptively navigate using real-time sensory data, bridging perception with action.
Yet, challenges persist, including generalization across scenarios, maintaining reliability, and ensuring real-time responsiveness. 
This paper introduces a perception framework grounded in foundation models for universal object detection and tracking, moving beyond specific training categories.
Integral to our approach is a multi-layered tracker integrated with the foundation detector, ensuring continuous target visibility, even when faced with motion blur, abrupt light shifts, and occlusions.
Complementing this, we introduce a model-free controller tailored for resilient quadrotor visual tracking.
Our system operates efficiently on limited hardware, relying solely on an onboard camera and an inertial measurement unit.
Through extensive validation in diverse challenging indoor and outdoor environments, we demonstrate our system's effectiveness and adaptability.
In conclusion, our research represents a step forward in quadrotor visual tracking, moving from task-specific methods to more versatile and adaptable operations.
\end{abstract}


\section*{Supplementary Material}
\noindent \textbf{Video}: \url{https://youtu.be/35sX9C1wUpA}

\section{Introduction} \label{sec:introduction}
Unmanned aerial vehicles, especially quadrotors, have recently proliferated across multiple applications like search and rescue, transportation, and inspections thanks to their agility, affordability, autonomy, and maneuverability~\cite{emran2018reviewquadrotor, rao2022quadformer, morando2022thermal, saviolo2023autocharge, lee2022vision, fan2019real, sikora2023towards}.
These advantages catalyzed research in control algorithms, with visual navigation as an emerging paradigm~\cite{courbon2009visual, zheng2015robust, scaramuzza2022learning}.

Historically, quadrotor control relied on trajectory-tracking methods from predetermined waypoints~\cite{bouabdallah2007full, eilers2020underactuated, quan2023robust, saviolo2022pitcn, cioffi2023hdvio}.
Although rooted in physics, these methods posed challenges due to the transformation of sensory data through multiple abstraction levels.
This often led to computational burdens and latency, especially in dynamic conditions~\cite{guo2020image, saviolo2023learning, yeom2023geometric, saviolo2022activelearningdynamics}.

Addressing these challenges, visual control methods have emerged, unifying perception and control. 
By processing real-time sensory data directly for robotic control, these methods offer improved performance in dynamic settings~\cite{zheng2018ibvs, manuelli2020keypoints, han2021fast, pan2021fast}. 
However, their initial dependency on hand-designed features and rigorous calibration made them less effective in unpredictable settings~\cite{espiau1992new, darma2013visual, chaumette2016visual, lin2022robust}.

The advent of deep learning has dramatically improved this area, making it more precise and adaptable~\cite{qin2023perception, wu2023daydreamer, loquercio2021learning, saxena2017exploring, shirzadeh2017vision, kulkarni2019unsupervised, sadeghi2019divis, rajguru2020camera}. 
However, ensuring consistent performance across diverse environments remains challenging.
The segment-anything model (SAM) shows promise with its zero-shot generalization~\cite{kirillov2023segment}.
However, the real-time inference capability of this \textit{foundation model} limits its usability in robotics, a limitation that persists even in its recent adaptations~\cite{zhao2023fast, yang2023track, rajivc2023segment}.

Our research delves into visual control for quadrotors, focusing on controlling the robot to detect, track, and follow arbitrary targets with ambiguous intent in challenging conditions.
We chose this task to validate the real-time data processing and generalization of this technique.
Effective target tracking under complex flight conditions requires real-time sensory data processing and robust generalization across scenarios.
Our contributions are threefold:
(i) Development of a perception framework optimized for real-time monocular detection and tracking. Our approach integrates foundation models for detecting and tracking beyond predefined categories. We introduce a tracker utilizing spatial, temporal, and appearance data to ensure continuous target visibility, even when faced with motion blur, abrupt light shifts, and occlusions.
(ii) Introduction of a model-free controller for quadrotor visual tracking that ensures the target remains in the camera's view while minimizing the robot's distance from the target.
Our controller relies solely on an onboard camera and inertial measurement unit, providing effective operation in GPS-denied environments.
(iii) Comprehensive validation of our system across varied indoor and outdoor environments, emphasizing its robustness and adaptability.

\section{Related Works} \label{sec:related_works}
\textbf{Detection and Tracking}.
Historically, object detection and tracking predominantly depended on single-frame detections~\cite{zhang2017real, braso2020learning, jung2022improved, reis2023real}.
Within this paradigm, unique features were extracted from bounding boxes in isolated frames to identify and track objects across temporal video sequences consistently. 
Among them, YOLO-based models were the most popular detectors adopted, thanks to their exceptional accuracy and efficiency~\cite{bolya2019yolact, song2021object, terven2023comprehensive, zhang2021toward}. 

A recent innovation is the SAM model~\cite{kirillov2023segment}, which is recognized for its superior generalization capability. 
Although its capabilities were unparalleled, SAM's architecture is inherently computationally demanding, making it suboptimal for real-time applications.
This drove the development of the optimized FastSAM~\cite{zhao2023fast}, which integrated YOLO to offset SAM's computational constraints.

Despite the advantages of these methods, they faced significant challenges with rapid object motion and occlusions, resulting in trajectory prediction inaccuracies.
Therefore, several studies leveraged tracking algorithms' temporal data to enhance object detection~\cite{aharon2022bot, du2023strongsort, jadhav2020aerial}.
For instance, hybrid approaches such as TAM~\cite{yang2023track} and SAM-PT~\cite{rajivc2023segment} were developed by combining the high-performance SAM detector with state-of-the-art trackers such as XMem and PIPS~\cite{cheng2022xmem, harley2022particle}.
Though effective, these approaches compromised real-time processing, a critical attribute for deployment on embedded platforms such as quadrotors.

\textbf{Visual Control}.
Visual control refers to the use of visual feedback to control the motion of robots~\cite{thomas2014toward, pan2021fast}. 
Historically, advancements in visual control for robotics were linked to the principles of image-based visual servoing (IBVS)~\cite{hutchinson1996tutorial}.
IBVS generated control commands for the robot by evaluating the differences between the current and a reference image.
Early results in IBVS showed promise in achieving precise control in various robotic applications~\cite{malis2001hybrid, chaumette2004recent}.
Specifically for quadrotors, \cite{azrad2010visual, bourquardez2009image, falanga2017vision} introduced IBVS methods for target tracking and maintaining positions over landmarks like landing pads.

Recent research in aerial vehicle visual control focuses on target tracking in dynamic 3D environments.
Multiple studies investigate model-based and robust visual control~\cite{roque2020fast, sheng2019image, zhao2019robust, yi2023neural, leomanni2023robust} and develop control policies using imitation and reinforcement~\cite{sampedro2018image, bhagat2020uav, fu2023deep, gervet2023navigating, kaiser2019model, dugas2022navdreams, brohan2022rt, brohan2023rt}.
For instance, \cite{guo2020image} proposes a visual control method that enables a quadrotor to fly through multiple openings at high speeds.
Concurrently, \cite{keipour2022visual} and \cite{cho2022autonomous} demonstrate methods for landing on moving platforms.
Recently, \cite{wang2023image} presents a visual control mechanism tailored for tracking arbitrary aerial targets in 3D settings.

Yet, many of these techniques presuppose specific knowledge about the system, targets, or environmental conditions, which restricts their broader adaptability. 
In contrast, our methodology abstains from modeling assumptions. 
Through the integration of foundation vision models, we maximize generalization, ensuring our system's adaptability and efficacy across diverse scenarios, and bolstering its applicability and robustness in real-world settings.

\begin{figure*}[t]
    \centering
    \includegraphics[width=\textwidth, trim=33 230 40 10, clip]{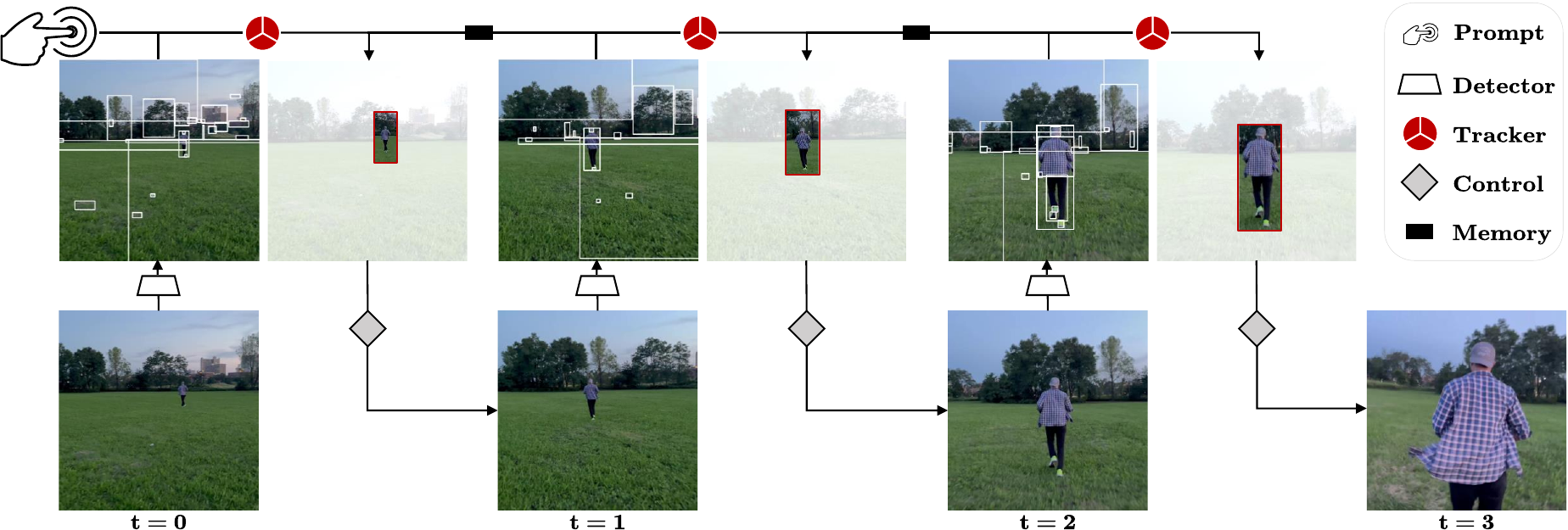}
    \caption{
    Our proposed framework for detecting (white), tracking (red), and following arbitrary targets using quadrotors. 
    The user-prompted target is detected and tracked over time by combining a real-time foundation detector with our novel multi-layered tracker.
    The quadrotor is continually controlled to navigate toward the target while maintaining it in the robot's view.
    \label{fig:methodology}}
        \vspace{-20pt}
\end{figure*}

\section{Methodology} \label{sec:methodology}
The objective is to enable a quadrotor to detect, track, and follow an arbitrary target with ambiguous intent in challenging conditions (Figure~\ref{fig:methodology}).
We tackle this by integrating:
(i) a universal detector with real-time accuracy. Unlike traditional models, ours harnesses foundation models, detecting objects beyond predefined categories;
(ii) a tracker to synergize with our foundation detector. It leverages spatial, temporal, and appearance data, ensuring target visibility despite challenges like motion blur, sudden lighting alterations, and occlusions;
(iii) a model-free visual controller for quadrotor tracking. This optimizes the target's presence within the camera's field while minimizing distance from the target.

\subsection{Target-Agnostic Real-Time Detection}
Ensuring model generalization on unseen data is essential in diverse, real-world settings.
While conventional models excel in specific conditions, they might struggle in unpredictable environments. 
Foundation models, with their vast datasets, cover a wider array of feature representations.

Building on SAM~\cite{kirillov2023segment} and FastSAM~\cite{zhao2023fast}, we utilize the YOLACT architecture~\cite{bolya2019yolact} integrated with foundation model strengths.
The detector employs a ResNet-101 backbone with a feature pyramid network for multi-scale feature maps.
For an image sized $(H, W, 3)$, it produces bounding boxes $B = { b_1, \ldots, b_n }$, where $b_i = (x_i, y_i, w_i, h_i)$ denotes the bounding box’s top-left coordinate, width, and height.

\subsection{Multi-layered Tracking}
\textbf{Initialization}.
To initiate tracking at time $t = 0$, a user identifies a point $(p_x, p_y)$ on the image to signify the target.
The L2 distance between this point and the center of each bounding box is calculated as
\begin{equation}
    d_i =
    \Vert
    ( p_x, p_y ) -
    ( x_i + w_i/2, y_i + h_i/2 )
    \Vert_2 .
\end{equation}
The bounding box closest to this point is the initial target
\begin{equation}
    b_{\text{target}}^{t=0} = \arg\min_{b_i^{t} \in B} d_i.
\end{equation}
Once initialized, tracking $b_{\text{target}}^{t=0}$ amid challenges like motion blurs, occlusions, and sensor noise is vital.

\textbf{Tracking}.
The objective is selecting the optimal bounding box encapsulating the target from the detector's outputs at each time $t$.
This is defined as a maximization problem
\begin{equation} \label{eq:tracking}
    b_{\text{target}}^{t} = \arg\max_{b_i^{t} \in B} s_{\text{target}}\left(b_i^{t}\right),
\end{equation}
where the combined score for bounding box $b_i^{t}$ is
\begin{equation}
    \begin{split}
        s_{\text{target}}\left(b_i^{t}\right) 
        &= \lambda_{\text{IOU}} s_{\text{IOU}}\left(b_i^{t}\right) \\
        &+ \lambda_{\text{EKF}} s_{\text{EKF}}\left(b_i^{t}\right) \\
        &+ \lambda_{\text{MAP}} s_{\text{MAP}}\left(b_i^{t}\right),
    \end{split}
\end{equation}
where scores $s_{\text{IOU}}$, $s_{\text{EKF}}$, and $s_{\text{MAP}}$ represent spatial coherence (intersection over union, IOU), temporal prediction via an extended Kalman filter (EKF), and appearance stability (through cosine similarity of multi-scale feature maps, MAP), respectively.
The associated weights, $\lambda_{\text{IOU}}$, $\lambda_{\text{EKF}}$, and $\lambda_{\text{MAP}}$, determine each metric's priority.

\subsubsection{Spatial Coherence}
Consistent correspondence between the target's bounding box at consecutive frames is essential. The IOU metric measures this consistency. For bounding boxes $b_1$ and $b_2$, the IOU is
\begin{equation}
\text{IOU}\left(b_1, b_2\right) = \frac{\text{area}(b_1 \cap b_2)}{\text{area}\left(b_1 \cup b_2\right)},
\end{equation}
where $\cap$ and $\cup$ denote intersection and union, and \text{area} of a bounding box is its width times height.
For each bounding box $b_i^{t}$ in the frame, its IOU score to the last target bounding box is
\begin{equation}
s_{\text{IOU}}(b_i^{t}) =
\text{IOU}\left(b_{\text{target}}^{t-1}, b_i^{t}\right).
\end{equation}

\subsubsection{Temporal Consistency}
Spatial coherence can be insufficient, especially with occlusions or inconsistencies. To enhance tracking, we combine spatial data with kinematic information using an EKF with a constant velocity model and gyroscope compensation.

Given an image sized $(H, W, 3)$ with target's bounding box $b_\text{target} = (x, y, w, h)$, and gyroscope's angular velocity in the camera frame, $\begin{bmatrix}\omega_x & \omega_y & \omega_z\end{bmatrix}^\top$, we define state and control input vectors as
\begin{equation}
\arraycolsep=2.25pt
\mathbf{x} =
\begin{bmatrix}
x & y & w & h & \dot{x} & \dot{y}
\end{bmatrix}^\top
\quad \text{and} \quad
\mathbf{u} =
\begin{bmatrix}
\omega_x & \omega_y & \omega_z
\end{bmatrix}^\top.
\end{equation}

The target's motion and covariance are modeled through a nonlinear stochastic equation, with constant velocity and angular velocity compensation to be robust to fast camera rotations. For details, the reader can refer to~\cite{ge2022vision}. Our model augments states $w, h$, consistent in the process model.

Each gyroscope measurement lets the EKF predict $\hat{b}_{\text{target}}$ for the visual controller. This operation lets the controller function at gyroscope speed, independent of camera and detector rates.

Upon receiving a frame, each bounding box $b_i^{t}$ receives an EKF score, computed as
\begin{equation}
s_{\text{EKF}}(b_i^{t}) =
\text{IOU}\left(\hat{b}_{\text{target}}, b_i^{t}\right).
\end{equation}

Following the optimization in eq.~(\ref{eq:tracking}), the EKF updates state and covariance vectors as in \cite{ge2022vision}.



\subsubsection{Appearance Robustness with Memory}
In tracking with unlabeled bounding boxes across frames, the object's appearance is crucial for consistency.
Feature maps from the detector capture object variations as continuous descriptors.
These maps help re-identify objects during occlusions or detector inconsistencies.

For each bounding box, our tracker extracts appearance information from feature maps. Each box is associated with feature vectors from the detector's features at its center.

To enhance appearance tracking, we integrate a memory mechanism using a complementary filter (CF).
At each iteration, the tracker computes the appearance score, $s_{\text{MAP}}\left(b_i^{t}\right)$, as the cosine similarity between the CF memory-stored features and the features of the current bounding box
\begin{equation}
s_{\text{MAP}}\left(b_i^{t}\right) =
\frac
{\mathbf{F}_\text{memory}^{t} \cdot \mathbf{F}_{b_i^{t}}}
{\Vert \mathbf{F}_\text{memory}^{t} \Vert_2 \times \Vert \mathbf{F}_{b_i^{t}} \Vert_2},
\end{equation}
where $\mathbf{F}_{\text{memory}}^{t}$ and $\mathbf{F}_{b_i^{t}}$ are the feature vectors from the CF's memory and the current bounding box.
The CF memory update is
\begin{equation}
\mathbf{F}_{\text{memory}}^{t} \leftarrow
\alpha \mathbf{F}_{\text{memory}}^{t-1} +
\left(1 - \alpha\right)\mathbf{F}_{\text{target}}^{t},
\end{equation}
where $ \alpha \in [0, 1] $ and $\mathbf{F}_{\text{target}}^{t}$ is the feature from the predicted target bounding box.
To avoid biases, we obscure areas outside the target box, re-feed this masked image to the detector, and re-extract the center feature values.
This ensures that the retained memory mainly comes from the target.
This operation is batch-parallelized, not affecting the detector's inference time.




\subsection{Visual Control}
Inspired by admittance control and $SE(3)$ group theory, we formulate a model-free visual controller that processes raw data from the camera and inertial sensors to reduce its distance to the target while keeping it in its field of view.

In the dynamics of quadrotors, forward movement results in pitching. 
This pitching action can change the target's position on the camera plane. 
Notably, the pitch predominantly influences the vertical position of the target, emphasizing the need to center the target in the image for optimal tracking. 
To account for this, the vertical setpoint $s_y$ is adjusted by the current pitch angle, $\theta$.
For an image of dimensions $(H, W, 3)$, the setpoints are
\begin{equation}
    s_x = \frac{W}{2}  
    \quad \text{and} \quad
    s_y = \frac{H}{2} - 2\frac{\theta}{\textit{v}} ,
\end{equation}
with $\textit{v}$ as the vertical field of view obtained through camera calibration~\cite{szeliski2022computer}.
%
%
Then, the position errors between the setpoint $(s_x, s_y)$ and target's predicted location $(p_x, p_y)$ are
\begin{equation}
    e_{w} = s_x - p_x
    \quad \text{and} \quad
    e_{h} = s_y - p_y .
\end{equation}

These errors drive the desired force in the world frame
\begin{equation}
    \arraycolsep=1.25pt
    \boldsymbol{f}_d ^{t} = 
    m \left(
    \mathbf{R}
    \begin{bmatrix}
        \hat{a}_{\theta} ^{t} \\
        k_{p\phi} e_{w} + k_{d\phi} \dot{e}_{w} \\
        k_{p\tau} e_{h} + k_{d\tau} \dot{e}_{h}
    \end{bmatrix} + 
    \mathbf{g}
    \right),
\end{equation}
where $k_{p\phi}$, $k_{d\phi}$, $k_{p\tau}$, $k_{d\tau}$ are respectively proportional and derivative gains for roll and thrust, $m$ is the quadrotor's mass, $\mathbf{R}$ is the current rotation of the quadrotor obtained by the inertial measurement unit, and $\mathbf{g} = \arraycolsep=2pt \begin{bmatrix} 0 & 0 & -9.81 \end{bmatrix}^\top$ represents the gravity vector.
The desired pitch acceleration at each step is determined by
\begin{equation}
    \hat{a}_{\theta} ^{t} = 
    \beta \hat{a}_{\theta} ^{t-1} +
    (1 - \beta) \bar{a}_{\theta},
\end{equation}
with $\beta \in [0, 1]$ modulating rise time and $\bar{a}_{\theta}$ representing a fixed pitch acceleration.
The CF is essential for limiting quadrotor jerks during initial tracking phases.

Through these equations, the controller derives the desired force based on the error between the detector's prediction and the setpoints. 
Subsequently, using $\arraycolsep=1.25pt \mathbf{e}_3 = \begin{bmatrix} 0 & 0 & 1 \end{bmatrix}^\top$, this force is converted into thrust
\begin{equation}
    \tau_d = (\mathbf{R}^{-1} \mathbf{f}_d)^\top \mathbf{e}_3.
\end{equation}

To find the desired orientation, we first compute the desired yaw $\psi_d$ using
\begin{equation}
    \psi_d = 
    \psi + 
    ( k_{p\psi} e_{w} + k_{d\psi} \dot{e}_{w} ) 
    \delta t,
\end{equation}
with $\psi$ as the quadrotor's current yaw and $k_{p\psi}$ and $k_{d\psi}$ proportional and derivative gains for yaw.
Then, the desired orientation matrix $\mathbf{R}_d$ is
\begin{equation}
    \mathbf{R}_d = 
    \arraycolsep=2pt 
    \begin{bmatrix} 
        \mathbf{r}_{1} & \mathbf{r}_{2} & \mathbf{r}_{3} 
    \end{bmatrix},
\end{equation}
with column vectors (following the ZYX convention)
\begin{equation}
    \begin{aligned}
        \mathbf{r}_{3} & = 
        \boldsymbol{f}_d / \Vert \boldsymbol{f}_d \Vert, \\
        \mathbf{r}_{2} & = 
        \begin{bmatrix} cos(\psi_d) & sin(\psi_d) & 0 \end{bmatrix}^\top, \\
        \mathbf{r}_{1} & = 
        \mathbf{r}_{2} \times \mathbf{r}_{3}.
    \end{aligned}
\end{equation}

Finally, the controller's thrust and orientation predictions guide an attitude PID controller in generating motor commands.
For the mathematical formulation of the PID controller~\cite{loianno2016estimation}.



\section{Experimental Results} \label{sec:exp_results}
Our evaluation procedure addresses the following questions:
(i) How does our perception framework generalize to common and rare object categories?
(ii) Is the tracker resilient to occlusions and detection failures?
(iii) What is the dependency of the tracker parameters?
(iv) How reliable is the approach across challenging flight conditions?
For a comprehensive understanding, we encourage the reader to consult as well the attached multimedia material.

\subsection{Setup} \label{sec:exp_setup}
\textbf{System}.
Our quadrotor, with a mass of $1.3~\si{kg}$, is powered by a $6$S battery and four motors.
It uses an NVIDIA Jetson Xavier NX for processing and captures visual data through an Arducam IMX477 at $544 \times 960$.
The framework operates in real-time: the controller processes at $100~\si{Hz}$ with each inertial measurement, while the perception framework is at $60~\si{Hz}$ with each camera frame.
Tests occurred at New York University in a large indoor environment measuring $26\times10\times4~\si{m^3}$, situated at the Agile Robotics and Perception Lab (ARPL) and in large outdoor fields in New York State.

\textbf{Perception}.
We utilize the Ultralytics repository~\cite{Jocher_YOLO_by_Ultralytics_2023} for implementing the foundation detector, modifying the code to extract feature maps at scales of $1/4$, $1/8$, $1/16$, and $1/32$, and efficiently exporting the model through TensorRT~\cite{tensorrt}.
Our designed tracker uses weights $\lambda_{\text{IOU}}=3$, $\lambda_{\text{EKF}}=3$, $\lambda_{\text{MAP}}=4$, CF rise time $\alpha=0.9$, and EKF process and measurement noise matrices respectively $\mathbf{Q} = \text{diag}(\begin{bmatrix} 0.01&0.01&0.01&0.01&0.1&0.1 \end{bmatrix})$ and $\mathbf{R} = \text{diag}(\begin{bmatrix} 0.5&0.5&0.5&0.5 \end{bmatrix})$.

\textbf{Control}.
For target distance, indoor and outdoor quadrotor pitch accelerations are $\bar{a}_{\theta} = 0.5 ~\si{m/s^2}$ and $\bar{a}_{\theta} = 5 ~\si{m/s^2}$.
The controller uses gains $k_{p\phi} = 0.05$, $k_{d\phi} = 0.001$, $k_{p\tau} = 0.08$, $k_{d\tau} = 0.00025$, $k_{p\psi} = 0.095$, $k_{d\psi} = 0.0004$, and $\beta=0.15$.

\begin{figure*}
    \begin{minipage}{\columnwidth}
        \setlength\tabcolsep{4.5pt}
        \centering
        \captionof{table}{\label{tab:generalization_common}}
        \vspace{-0.75em}
        \caption*{\scshape Tracking Common Object Categories}
        \begin{tabular}{c c c c c c}
            \toprule\toprule
            Category & \# Frames & IOU $[\%]$ & Overlap $[\%]$ & Tracked $[\%]$ \\
            \midrule
            Human      & $1763$ & $42$ & $91$ & $100$ \\
            Car        & $2120$ & $54$ & $95$ & $100$ \\
            Tv         & $890$  & $90$ & $98$ & $100$ \\
            Table      & $2304$ & $92$ & $97$ & $100$ \\
            Umbrella   & $791$  & $31$ & $88$ & $100$ \\
            \bottomrule\bottomrule
        \end{tabular}
    \end{minipage}
    \hfill
    \begin{minipage}{\columnwidth}
        \setlength\tabcolsep{4.5pt}
        \centering
        \captionof{table}{\label{tab:ablation_study}}
        \vspace{-0.75em}
        \caption*{\scshape Tracker Ablation Study}
        \begin{tabular}{c c c c c c c}
            \toprule\toprule
            \multirow{2}{*}{$\lambda_{\text{IOU}}$} & 
            \multirow{2}{*}{$\lambda_{\text{EKF}}$} & 
            \multirow{2}{*}{$\lambda_{\text{MAP}}$} && Human && Drone \\
            \cline{5-5} \cline{7-7}
            & & && Tracked $[\%]$ && Tracked $[\%]$\\
            \midrule
            3 & 0 & 0 && $8$ && $2$ \\
            3 & 3 & 0 && $49$ && $11$ \\
            3 & 0 & 4 && $100$ && $24$ \\
            3 & 3 & 4 && $100$ && $100$ \\
            \bottomrule\bottomrule
        \end{tabular}
    \end{minipage}
            \vspace{-5pt}
\end{figure*}

\begin{figure*}[t]
    \centering
    \includegraphics[width=\linewidth, trim=20 420 20 0, clip]
    {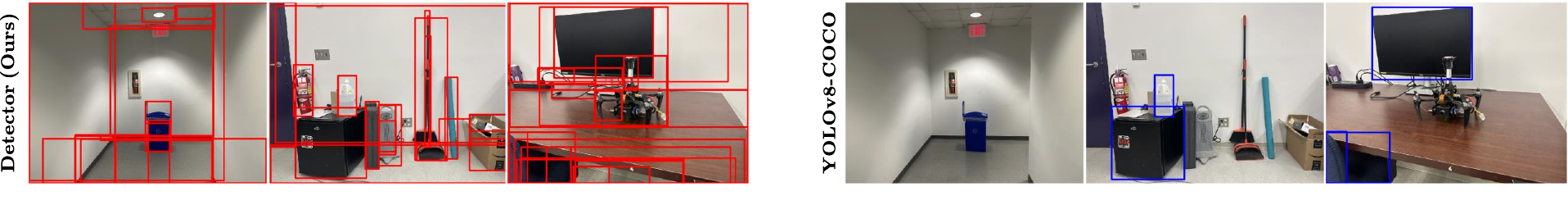}
    \vspace{-2em}
    \caption{
    Despite the YOLO baseline's extensive training on 80 categories, it struggles to recognize custom drones, irregular trash cans, and pool noodles. In contrast, our foundation detector showcases significant adaptability and robustness, accurately identifying these unique objects without prior specific training.
    }
    \label{fig:generalization_rare}
        \vspace{-10pt}
\end{figure*}

\begin{figure*}[t]
    \centering
    \includegraphics[width=\linewidth, trim=75 252 75 0, clip]{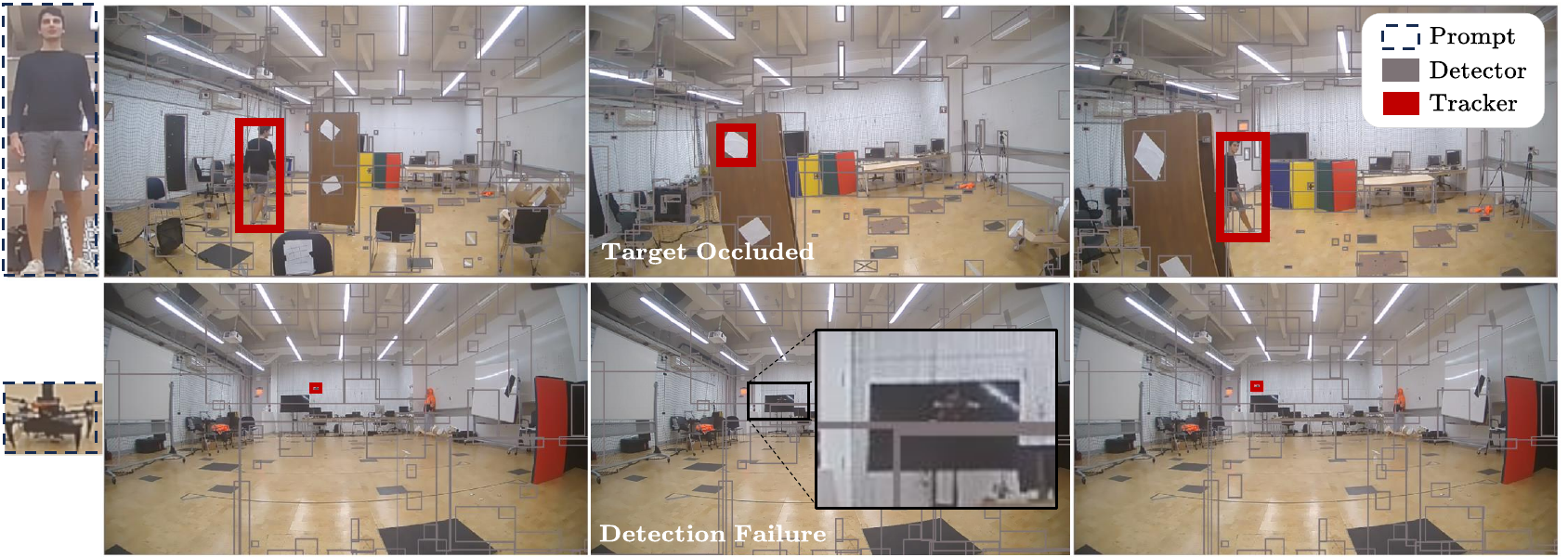}
    \caption{
    Our detection and tracking algorithm's resilience and accuracy.
    Top row: Indoor spatial-temporal tracking of a human against occlusions. 
    Bottom row: Tracking of our custom-made drone, highlighting re-identification capabilities.
    }
    \label{fig:tracking_disruption}
    \vspace{-15pt}
\end{figure*}

\subsection{Perception Generalization Performance}

\textbf{Common Categories}.
We test our perception framework with drone-captured videos of common objects~\cite{lin2014microsoft} and compare it to a YOLO baseline model~\cite{Jocher_YOLO_by_Ultralytics_2023}.
We initiate our tracker using the baseline's first detected object.
Each indoor video features a stationary target, including TV, table, and umbrella, against cluttered backgrounds.
Outdoor videos capture moving targets like a car on an asphalt road or a human running in a park.
Our drone maintains its tracking with no target occlusions.
For evaluation, we employ several metrics.
The \textit{IOU} assesses the overlap between our predictions and the YOLO baseline.
\textit{Overlap} measures how often our model's predictions align with the baseline's predictions,
\textit{Tracked} represents the proportion of frames where the target was consistently tracked to the video's conclusion.

The results, detailed in Table~\ref{tab:generalization_common}, indicate that our model tracks closely to the YOLO baseline with an overlapping frequency of about $94\%$.
However, the IOU score is roughly $62\%$.
This discrepancy is attributed to our foundation model's non-target-aware nature during its training phase. 
As a result, it occasionally recognizes multiple bounding boxes for a singular target, which impacts the IOU score negatively.

\textbf{Rare Categories}.
We evaluate the generalization capabilities of our perception framework to rare categories of objects, including custom-made drones, irregularly shaped trash cans, and soft pool noodles.
The results, illustrated in Figure~\ref{fig:generalization_rare}, qualitatively compare the performance of our foundation detector with the YOLO baseline.
Even though the baseline is comprehensively trained on $80$ categories, the rare characteristics and shapes of the considered objects make the baseline fail in detecting the instances in the scene.
On the contrary, our foundation detector demonstrates high generalization capabilities, successfully detecting and tracking these rare objects with consistent and remarkable accuracy.
Despite not being specifically trained for these categories, the inherent flexibility and adaptive nature of our perception framework allow it to generalize beyond its training set.
These results validate the adaptability to different and novel object categories of our model.

\begin{figure*}
    \centering
    \includegraphics[width=\linewidth, trim=0 340 0 40, clip]{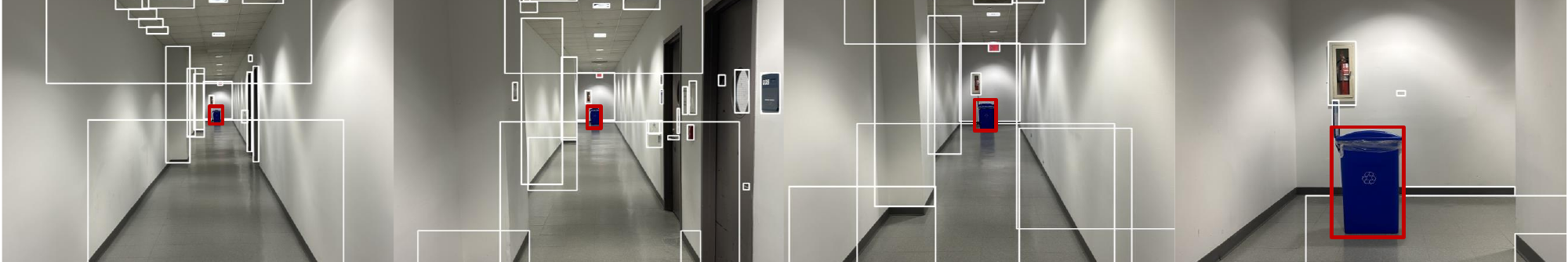}
    \caption{
    Demonstrating our system's ability to detect, track, and navigate toward an asymmetrically shaped trash can within a featureless narrow corridor, hence highlighting the efficacy of our visual control method.
    }
    \label{fig:trash_tracking}
            \vspace{-10pt}
\end{figure*}
\begin{figure*}
    \centering
    \includegraphics[width=\linewidth, trim=0 360 0 20, clip]{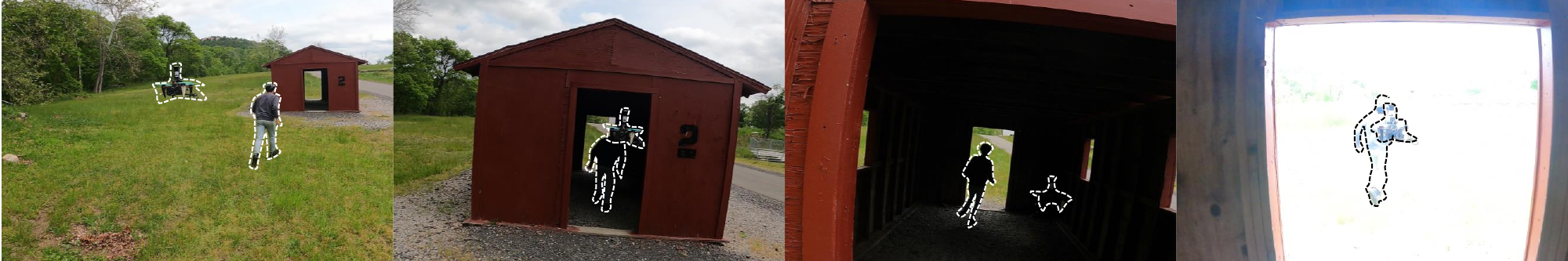}
    \caption{
    The human target transitions from outdoor to indoor settings, showcasing our system's ability to adapt to rapid lighting shifts and maintain consistent tracking.
    }
    \label{fig:through_house}
        \vspace{-10pt}
\end{figure*}

\subsection{Tracking Resiliency Post-Disruption}
Our comprehensive analysis further investigates the re-identification ability of our approach after potential detection failure and target occlusions. 
As shown in Figure~\ref{fig:tracking_disruption}, we conduct our experiments in a controlled indoor setting.
Both the camera and the targets are actively moving, and there are multiple instances where the target is strategically occluded.

First, we track a human target. 
The results show that our system can effectively track even when the target is briefly occluded, without needing specific category training.

Then, we track a custom-made drone in the same complex indoor environment. 
Due to its distinct irregular shape, the drone had a notably higher rate of false positives compared to the human.
However, our approach still manages to track the drone with accuracy, re-identifying it even after brief periods of misclassification, thanks to the deep feature similarity embedded in our tracker.

\subsection{Tracker Ablation Study}
We conduct an ablation study to evaluate each weight in our perception framework, particularly in conditions of indoor clutter and target occlusions.
For these experiments, we use human and drone video data from previous experiments, as illustrated in Figure~\ref{fig:tracking_disruption}.
The results are reported in Table~\ref{tab:ablation_study}.

In the human dataset, relying on spatial data is inadequate as occlusions or failed detections cause tracking failure.
Even adding temporal consistency does not compensate due to the target's unpredictable motion.
Prioritizing appearance robustness alone ensures tracking by facilitating re-identification after uncertain motions and occlusions, but the best accuracy is achieved when all components are combined.

For the drone dataset, depending solely on spatial coherence or combining it with temporal consistency is ineffective due to false positive detections.
Emphasizing only appearance robustness is also ineffective.
Because it depends on the smooth motion of the quadrotor, a feature only temporal consistency provides.
Thus, successful tracking for the drone dataset is achieved when all mechanisms are integrated.

\subsection{Flight Through Narrow Featureless Corridors}
We assess our system's ability to detect, track, and navigate toward an asymmetric trash can within a featureless corridor (Figure~\ref{fig:trash_tracking}).
During the flight, our system is capable of reaching speeds up to $4~\si{m/s}$.
This task demonstrates the superiority of our visual control methodology over traditional strategies. 
Conventional control methods would falter in such corridors due to their reliance on global state estimations that require unique landmarks. 
Conversely, our method leverages relative image-based localization, foregoing the need for global positional data. 
This boosts quicker adaptive responses and resilience in featureless settings. 
The illustrated results validate the system's capability to accurately track the target in this challenging environment, underscoring the prowess of visual control.

\subsection{Human Tracking Through Forests and Buildings}
We tested our approach in diverse outdoor settings, tracking a human target while sprinting on a road, moving through a building, and running in a dense forest (see Figure~\ref{fig:initial_figure} and Figure~\ref{fig:through_house}), with target speeds up to $7~\si{m/s}$.
Humans were chosen due to their unpredictable movement and no battery dependencies.
Transitions, such as entering a building, introduce lighting changes, requiring quick adaptation. The forest environment presented issues with foliage and varied light. Despite these challenges, our quadrotor consistently tracked the human, showcasing our system's adaptability and robustness.

\section{Conclusion and Future Works} \label{sec:conclusion}
Although promising, visual control faces challenges in generalization and reliability.
Addressing this, we introduced a visual control methodology optimized for target tracking under challenging flight conditions.
Central to our innovation are foundation vision models for target detection.
Our breakthrough hinges on our novel tracker and model-free controller, designed to integrate seamlessly with the foundation model.
Through experimentation in diverse environments, we showed that our approach not only addresses the challenges of visual control but also advances target tracking, moving from task-specific methods to versatile operations.

Moving forward, leveraging the detector's dense predictions will allow us to learn our robot's dynamics with respect to the target and its environment. 
To ensure safe navigation, we aim to incorporate a learning-based reactive policy in the controller, mitigating collision risks while maintaining target tracking. 
Furthermore, it is crucial to guarantee that the target remains within the camera's view. 
As such, we intend to explore predictive control solutions using control barrier functions. 
These enhancements will optimize our system's performance, fortifying its capability for intricate real-world applications.

\clearpage

\bibliographystyle{IEEEtran}
\bibliography{references}

\begin{thebibliography}{10}
\providecommand{\url}[1]{#1}
\csname url@rmstyle\endcsname
\providecommand{\newblock}{\relax}
\providecommand{\bibinfo}[2]{#2}
\providecommand\BIBentrySTDinterwordspacing{\spaceskip=0pt\relax}
\providecommand\BIBentryALTinterwordstretchfactor{4}
\providecommand\BIBentryALTinterwordspacing{\spaceskip=\fontdimen2\font plus
\BIBentryALTinterwordstretchfactor\fontdimen3\font minus
  \fontdimen4\font\relax}
\providecommand\BIBforeignlanguage[2]{{%
\expandafter\ifx\csname l@#1\endcsname\relax
\typeout{** WARNING: IEEEtran.bst: No hyphenation pattern has been}%
\typeout{** loaded for the language `#1'. Using the pattern for}%
\typeout{** the default language instead.}%
\else
\language=\csname l@#1\endcsname
\fi
#2}}

\bibitem{emran2018reviewquadrotor}
B.~J. Emran and H.~Najjaran, ``A review of quadrotor: An underactuated
  mechanical system,'' \emph{Annual Reviews in Control}, vol.~46, pp. 165--180,
  2018.

\bibitem{rao2022quadformer}
P.~P. Rao, F.~Qiao, W.~Zhang, Y.~Xu, Y.~Deng, G.~Wu, and Q.~Zhang,
  ``Quadformer: Quadruple transformer for unsupervised domain adaptation in
  power line segmentation of aerial images,'' \emph{arXiv preprint
  arXiv:2211.16988}, 2022.

\bibitem{morando2022thermal}
L.~Morando, C.~T. Recchiuto, J.~Calla, P.~Scuteri, and A.~Sgorbissa, ``Thermal
  and visual tracking of photovoltaic plants for autonomous uav inspection,''
  \emph{Drones}, vol.~6, no.~11, p. 347, 2022.

\bibitem{saviolo2023autocharge}
A.~Saviolo, J.~Mao, R.~B. TMB, V.~Radhakrishnan, and G.~Loianno, ``Autocharge:
  Autonomous charging for perpetual quadrotor missions,'' in \emph{IEEE
  International Conference on Robotics and Automation (ICRA)}, 2023, pp.
  5400--5406.

\bibitem{lee2022vision}
E.~S. Lee, G.~Loianno, D.~Jayaraman, and V.~Kumar, ``Vision-based perimeter
  defense via multiview pose estimation,'' \emph{arXiv preprint
  arXiv:2209.12136}, 2022.

\bibitem{fan2019real}
R.~Fan, J.~Jiao, J.~Pan, H.~Huang, S.~Shen, and M.~Liu, ``Real-time dense
  stereo embedded in a uav for road inspection,'' in \emph{Proceedings of the
  IEEE/CVF Conference on Computer Vision and Pattern Recognition Workshops}, 06
  2019, pp. 535--543.

\bibitem{sikora2023towards}
T.~Sikora, L.~Markovic, and S.~Bogdan, ``Towards operating wind turbine
  inspections using a lidar-equipped uav,'' \emph{arXiv preprint
  arXiv:2306.14637}, 2023.

\bibitem{courbon2009visual}
J.~Courbon, Y.~Mezouar, N.~Guenard, and P.~Martinet, ``Visual navigation of a
  quadrotor aerial vehicle,'' in \emph{IEEE/RSJ International Conference on
  Intelligent Robots and Systems (IROS)}, 2009, pp. 5315--5320.

\bibitem{zheng2015robust}
W.~Zheng, F.~Zhou, and Z.~Wang, ``Robust and accurate monocular visual
  navigation combining imu for a quadrotor,'' \emph{IEEE/CAA Journal of
  Automatica Sinica}, vol.~2, no.~1, pp. 33--44, 2015.

\bibitem{scaramuzza2022learning}
D.~Scaramuzza and E.~Kaufmann, ``Learning agile, vision-based drone flight:
  From simulation to reality,'' in \emph{Robotics Research}, A.~Billard,
  T.~Asfour, and O.~Khatib, Eds.\hskip 1em plus 0.5em minus 0.4em\relax Cham:
  Springer Nature Switzerland, 2023, pp. 11--18.

\bibitem{bouabdallah2007full}
S.~Bouabdallah and R.~Siegwart, ``Full control of a quadrotor,'' in
  \emph{IEEE/RSJ international conference on intelligent robots and systems
  (IROS)}, 2007, pp. 153--158.

\bibitem{eilers2020underactuated}
C.~Eilers, J.~Eschmann, R.~Menzenbach, B.~Belousov, F.~Muratore, and J.~Peters,
  ``Underactuated waypoint trajectory optimization for light painting
  photography,'' in \emph{IEEE International Conference on Robotics and
  Automation (ICRA)}, 2020, pp. 1505--1510.

\bibitem{quan2023robust}
L.~Quan, L.~Yin, T.~Zhang, M.~Wang, R.~Wang, S.~Zhong, X.~Zhou, Y.~Cao, C.~Xu,
  and F.~Gao, ``Robust and efficient trajectory planning for formation flight
  in dense environments,'' \emph{IEEE Transactions on Robotics}, vol.~39,
  no.~6, pp. 4785--4804, 2023.

\bibitem{saviolo2022pitcn}
A.~Saviolo, G.~Li, and G.~Loianno, ``Physics-inspired temporal learning of
  quadrotor dynamics for accurate model predictive trajectory tracking,''
  \emph{IEEE Robotics and Automation Letters}, vol.~7, no.~4, pp.
  10\,256--10\,263, 2022.

\bibitem{cioffi2023hdvio}
G.~Cioffi, L.~Bauersfeld, and D.~Scaramuzza, ``{HDVIO: Improving Localization
  and Disturbance Estimation with Hybrid Dynamics VIO},'' in \emph{Proceedings
  of Robotics: Science and Systems (RSS)}, 2023.

\bibitem{guo2020image}
D.~Guo and K.~K. Leang, ``Image-based estimation, planning, and control for
  high-speed flying through multiple openings,'' \emph{The International
  Journal of Robotics Research}, vol.~39, no.~9, pp. 1122--1137, 2020.

\bibitem{saviolo2023learning}
A.~Saviolo and G.~Loianno, ``Learning quadrotor dynamics for precise, safe, and
  agile flight control,'' \emph{Annual Reviews in Control}, vol.~55, pp.
  45--60, 2023.

\bibitem{yeom2023geometric}
J.~Yeom, G.~Li, and G.~Loianno, ``Geometric fault-tolerant control of
  quadrotors in case of rotor failures: An attitude based comparative study,''
  in \emph{IEEE/RSJ International Conference on Intelligent Robots and Systems
  (IROS)}, 2023, pp. 4974--4980.

\bibitem{saviolo2022activelearningdynamics}
A.~Saviolo, J.~Frey, A.~Rathod, M.~Diehl, and G.~Loianno, ``Active learning of
  discrete-time dynamics for uncertainty-aware model predictive control,''
  \emph{IEEE Transactions on Robotics}, vol.~40, pp. 1273--1291, 2024.

\bibitem{zheng2018ibvs}
D.~Zheng, H.~Wang, W.~Chen, and Y.~Wang, ``Planning and tracking in image space
  for image-based visual servoing of a quadrotor,'' \emph{IEEE Transactions on
  Industrial Electronics}, vol.~65, no.~4, pp. 3376--3385, 2018.

\bibitem{manuelli2020keypoints}
L.~Manuelli, Y.~Li, P.~Florence, and R.~Tedrake, ``Keypoints into the future:
  Self-supervised correspondence in model-based reinforcement learning,'' in
  \emph{Proceedings of the 2020 Conference on Robot Learning (CoRL)}, ser.
  Proceedings of Machine Learning Research, J.~Kober, F.~Ramos, and C.~Tomlin,
  Eds., vol. 155.\hskip 1em plus 0.5em minus 0.4em\relax PMLR, 2021, pp.
  693--710.

\bibitem{han2021fast}
Z.~Han, R.~Zhang, N.~Pan, C.~Xu, and F.~Gao, ``Fast-tracker: A robust aerial
  system for tracking agile target in cluttered environments,'' in \emph{IEEE
  International Conference on Robotics and Automation (ICRA)}, 2021, pp.
  328--334.

\bibitem{pan2021fast}
N.~Pan, R.~Zhang, T.~Yang, C.~Cui, C.~Xu, and F.~Gao, ``Fast-tracker 2.0:
  Improving autonomy of aerial tracking with active vision and human location
  regression,'' \emph{IET Cyber-Systems and Robotics}, vol.~3, no.~4, pp.
  292--301, 2021.

\bibitem{espiau1992new}
B.~Espiau, F.~Chaumette, and P.~Rives, ``A new approach to visual servoing in
  robotics,'' \emph{IEEE Transactions on Robotics and Automation}, vol.~8,
  no.~3, pp. 313--326, 1992.

\bibitem{darma2013visual}
S.~Darma, J.~L. Buessler, G.~Hermann, J.~P. Urban, and B.~Kusumoputro, ``Visual
  servoing quadrotor control in autonomous target search,'' in \emph{IEEE 3rd
  International Conference on System Engineering and Technology}, 2013, pp.
  319--324.

\bibitem{chaumette2016visual}
F.~Chaumette, S.~Hutchinson, and P.~Corke, ``Visual servoing,'' \emph{Springer
  handbook of robotics}, pp. 841--866, 2016.

\bibitem{lin2022robust}
J.~Lin, Y.~Wang, Z.~Miao, S.~Fan, and H.~Wang, ``Robust observer-based visual
  servo control for quadrotors tracking unknown moving targets,''
  \emph{IEEE/ASME Transactions on Mechatronics}, 2022.

\bibitem{qin2023perception}
C.~Qin, Q.~Yu, H.~S.~H. Go, and H.~H.-T. Liu, ``Perception-aware image-based
  visual servoing of aggressive quadrotor uavs,'' \emph{IEEE/ASME Transactions
  on Mechatronics}, vol.~28, no.~4, pp. 2020--2028, 2023.

\bibitem{wu2023daydreamer}
P.~Wu, A.~Escontrela, D.~Hafner, P.~Abbeel, and K.~Goldberg, ``Daydreamer:
  World models for physical robot learning,'' in \emph{Conference on Robot
  Learning (CoRL)}.\hskip 1em plus 0.5em minus 0.4em\relax PMLR, 2023, pp.
  2226--2240.

\bibitem{loquercio2021learning}
A.~Loquercio, E.~Kaufmann, R.~Ranftl, M.~M{\"u}ller, V.~Koltun, and
  D.~Scaramuzza, ``Learning high-speed flight in the wild,'' \emph{Science
  Robotics}, vol.~6, no.~59, p. eabg5810, 2021.

\bibitem{saxena2017exploring}
A.~Saxena, H.~Pandya, G.~Kumar, A.~Gaud, and K.~M. Krishna, ``Exploring
  convolutional networks for end-to-end visual servoing,'' in \emph{IEEE
  International Conference on Robotics and Automation (ICRA)}, 2017, pp.
  3817--3823.

\bibitem{shirzadeh2017vision}
M.~Shirzadeh, H.~J. Asl, A.~Amirkhani, and A.~A. Jalali, ``Vision-based control
  of a quadrotor utilizing artificial neural networks for tracking of moving
  targets,'' \emph{Engineering Applications of Artificial Intelligence},
  vol.~58, pp. 34--48, 2017.

\bibitem{kulkarni2019unsupervised}
T.~D. Kulkarni, A.~Gupta, C.~Ionescu, S.~Borgeaud, M.~Reynolds, A.~Zisserman,
  and V.~Mnih, ``Unsupervised learning of object keypoints for perception and
  control,'' \emph{Advances in neural information processing systems}, vol.~32,
  2019.

\bibitem{sadeghi2019divis}
F.~Sadeghi, ``Divis: Domain invariant visual servoing for collision-free goal
  reaching,'' in \emph{Proceedings of Robotics: Science and Systems (RSS)},
  FreiburgimBreisgau, Germany, June 2019.

\bibitem{rajguru2020camera}
A.~Rajguru, C.~Collander, and W.~J. Beksi, ``Camera-based adaptive trajectory
  guidance via neural networks,'' in \emph{2020 6th International Conference on
  Mechatronics and Robotics Engineering (ICMRE)}, 2020, pp. 155--159.

\bibitem{kirillov2023segment}
A.~Kirillov, E.~Mintun, N.~Ravi, H.~Mao, C.~Rolland, L.~Gustafson, T.~Xiao,
  S.~Whitehead, A.~C. Berg, W.-Y. Lo, P.~Dollar, and R.~Girshick, ``Segment
  anything,'' in \emph{Proceedings of the IEEE/CVF International Conference on
  Computer Vision (ICCV)}, October 2023, pp. 4015--4026.

\bibitem{zhao2023fast}
X.~Zhao, W.~Ding, Y.~An, Y.~Du, T.~Yu, M.~Li, M.~Tang, and J.~Wang, ``Fast
  segment anything,'' \emph{arXiv preprint arXiv:2306.12156}, 2023.

\bibitem{yang2023track}
J.~Yang, M.~Gao, Z.~Li, S.~Gao, F.~Wang, and F.~Zheng, ``Track anything:
  Segment anything meets videos,'' \emph{arXiv preprint arXiv:2304.11968},
  2023.

\bibitem{rajivc2023segment}
F.~Raji{\v{c}}, L.~Ke, Y.-W. Tai, C.-K. Tang, M.~Danelljan, and F.~Yu,
  ``Segment anything meets point tracking,'' \emph{arXiv preprint
  arXiv:2307.01197}, 2023.

\bibitem{zhang2017real}
Y.~Zhang, J.~Wang, and X.~Yang, ``Real-time vehicle detection and tracking in
  video based on faster r-cnn,'' in \emph{Journal of Physics: Conference
  Series}, vol. 887, no.~1.\hskip 1em plus 0.5em minus 0.4em\relax IOP
  Publishing, 2017, p. 012068.

\bibitem{braso2020learning}
G.~Bras{\'o} and L.~Leal-Taix{\'e}, ``Learning a neural solver for multiple
  object tracking,'' in \emph{IEEE/CVF International Conference on Computer
  Vision and Pattern Recognition (CVPR)}, 2020, pp. 6247--6257.

\bibitem{jung2022improved}
H.-K. Jung and G.-S. Choi, ``Improved yolov5: Efficient object detection using
  drone images under various conditions,'' \emph{Applied Sciences}, vol.~12,
  no.~14, p. 7255, 2022.

\bibitem{reis2023real}
D.~Reis, J.~Kupec, J.~Hong, and A.~Daoudi, ``Real-time flying object detection
  with yolov8,'' \emph{arXiv preprint arXiv:2305.09972}, 2023.

\bibitem{bolya2019yolact}
D.~Bolya, C.~Zhou, F.~Xiao, and Y.~J. Lee, ``Yolact: Real-time instance
  segmentation,'' in \emph{IEEE/CVF International Conference on Computer Vision
  (CVPR)}, 2019, pp. 9157--9166.

\bibitem{song2021object}
Q.~Song, S.~Li, Q.~Bai, J.~Yang, X.~Zhang, Z.~Li, and Z.~Duan, ``Object
  detection method for grasping robot based on improved yolov5,''
  \emph{Micromachines}, vol.~12, no.~11, p. 1273, 2021.

\bibitem{terven2023comprehensive}
J.~Terven, D.-M. Córdova-Esparza, and J.-A. Romero-González, ``A
  comprehensive review of yolo architectures in computer vision: From yolov1 to
  yolov8 and yolo-nas,'' \emph{Machine Learning and Knowledge Extraction},
  vol.~5, no.~4, pp. 1680--1716, 2023.

\bibitem{zhang2021toward}
T.~Zhang, J.~Xiao, L.~Li, C.~Wang, and G.~Xie, ``Toward coordination control of
  multiple fish-like robots: Real-time vision-based pose estimation and
  tracking via deep neural networks.'' \emph{IEEE CAA J. Autom. Sinica},
  vol.~8, no.~12, pp. 1964--1976, 2021.

\bibitem{aharon2022bot}
N.~Aharon, R.~Orfaig, and B.-Z. Bobrovsky, ``Bot-sort: Robust associations
  multi-pedestrian tracking,'' \emph{arXiv preprint arXiv:2206.14651}, 2022.

\bibitem{du2023strongsort}
Y.~Du, Z.~Zhao, Y.~Song, Y.~Zhao, F.~Su, T.~Gong, and H.~Meng, ``Strongsort:
  Make deepsort great again,'' \emph{IEEE Transactions on Multimedia}, pp.
  1--14, 2023.

\bibitem{jadhav2020aerial}
A.~Jadhav, P.~Mukherjee, V.~Kaushik, and B.~Lall, ``Aerial multi-object
  tracking by detection using deep association networks,'' in \emph{2020
  National Conference on Communications (NCC)}, 2020, pp. 1--6.

\bibitem{cheng2022xmem}
H.~K. Cheng and A.~G. Schwing, ``Xmem: Long-term video object segmentation with
  an atkinson-shiffrin memory model,'' in \emph{European Conference on Computer
  Vision}.\hskip 1em plus 0.5em minus 0.4em\relax Springer, 2022, pp. 640--658.

\bibitem{harley2022particle}
A.~W. Harley, Z.~Fang, and K.~Fragkiadaki, ``Particle video revisited: Tracking
  through occlusions using point trajectories,'' in \emph{European Conference
  on Computer Vision}.\hskip 1em plus 0.5em minus 0.4em\relax Springer, 2022,
  pp. 59--75.

\bibitem{thomas2014toward}
J.~Thomas, G.~Loianno, K.~Sreenath, and V.~Kumar, ``Toward image based visual
  servoing for aerial grasping and perching,'' in \emph{IEEE International
  Conference on Robotics and Automation (ICRA)}, 2014, pp. 2113--2118.

\bibitem{hutchinson1996tutorial}
S.~Hutchinson, G.~D. Hager, and P.~I. Corke, ``A tutorial on visual servo
  control,'' \emph{IEEE Transactions on Robotics and Automation}, vol.~12,
  no.~5, pp. 651--670, 1996.

\bibitem{malis2001hybrid}
E.~Malis, ``Hybrid vision-based robot control robust to large calibration
  errors on both intrinsic and extrinsic camera parameters,'' in \emph{European
  Control Conference (ECC)}, 2001, pp. 2898--2903.

\bibitem{chaumette2004recent}
F.~Chaumette and E.~Marchand, ``Recent results in visual servoing for robotics
  applications,'' in \emph{8th ESA Workshop on Advanced Space Technologies for
  Robotics and Automation (ASTRA)}, 2004, pp. 471--478.

\bibitem{azrad2010visual}
S.~Azrad, F.~Kendoul, and K.~Nonami, ``Visual servoing of quadrotor micro-air
  vehicle using color-based tracking algorithm,'' \emph{Journal of System
  Design and Dynamics}, vol.~4, no.~2, pp. 255--268, 2010.

\bibitem{bourquardez2009image}
O.~Bourquardez, R.~Mahony, N.~Guenard, F.~Chaumette, T.~Hamel, and L.~Eck,
  ``Image-based visual servo control of the translation kinematics of a
  quadrotor aerial vehicle,'' \emph{IEEE Transactions on Robotics}, vol.~25,
  no.~3, pp. 743--749, 2009.

\bibitem{falanga2017vision}
D.~Falanga, A.~Zanchettin, A.~Simovic, J.~Delmerico, and D.~Scaramuzza,
  ``Vision-based autonomous quadrotor landing on a moving platform,'' in
  \emph{IEEE International Symposium on Safety, Security and Rescue Robotics
  (SSRR)}, 2017, pp. 200--207.

\bibitem{roque2020fast}
P.~Roque, E.~Bin, P.~Miraldo, and D.~V. Dimarogonas, ``Fast model predictive
  image-based visual servoing for quadrotors,'' in \emph{IEEE/RSJ International
  Conference on Intelligent Robots and Systems (IROS)}, 2020, pp. 7566--7572.

\bibitem{sheng2019image}
H.~Sheng, E.~Shi, and K.~Zhang, ``Image-based visual servoing of a quadrotor
  with improved visibility using model predictive control,'' in \emph{IEEE 28th
  International Symposium on Industrial Electronics (ISIE)}, 2019, pp.
  551--556.

\bibitem{zhao2019robust}
W.~Zhao, H.~Liu, F.~L. Lewis, K.~P. Valavanis, and X.~Wang, ``Robust visual
  servoing control for ground target tracking of quadrotors,'' \emph{IEEE
  Transactions on Control Systems Technology}, vol.~28, no.~5, pp. 1980--1987,
  2019.

\bibitem{yi2023neural}
X.~Yi, B.~Luo, and Y.~Zhao, ``Neural network-based robust guaranteed cost
  control for image-based visual servoing of quadrotor,'' \emph{IEEE
  Transactions on Neural Networks and Learning Systems}, 2023.

\bibitem{leomanni2023robust}
M.~Leomanni, F.~Ferrante, N.~Cartocci, G.~Costante, M.~L. Fravolini, K.~M.
  Dogan, and T.~Yucelen, ``Robust output feedback control of a quadrotor uav
  for autonomous vision-based target tracking,'' in \emph{AIAA SCITECH 2023
  Forum}, 2023, p. 1632.

\bibitem{sampedro2018image}
C.~Sampedro, A.~Rodriguez-Ramos, I.~Gil, L.~Mejias, and P.~Campoy,
  ``Image-based visual servoing controller for multirotor aerial robots using
  deep reinforcement learning,'' in \emph{IEEE/RSJ International Conference on
  Intelligent Robots and Systems (IROS)}, 2018, pp. 979--986.

\bibitem{bhagat2020uav}
S.~Bhagat and P.~Sujit, ``Uav target tracking in urban environments using deep
  reinforcement learning,'' in \emph{International Conference on Unmanned
  Aircraft Systems (ICUAS)}, 2020, pp. 694--701.

\bibitem{fu2023deep}
G.~Fu, H.~Chu, L.~Liu, L.~Fang, and X.~Zhu, ``Deep reinforcement learning for
  the visual servoing control of uavs with fov constraint,'' \emph{Drones},
  vol.~7, no.~6, p. 375, 2023.

\bibitem{gervet2023navigating}
T.~Gervet, S.~Chintala, D.~Batra, J.~Malik, and D.~S. Chaplot, ``Navigating to
  objects in the real world,'' \emph{Science Robotics}, vol.~8, no.~79, p.
  eadf6991, 2023.

\bibitem{kaiser2019model}
L.~Kaiser, M.~Babaeizadeh, P.~Milos, B.~Osinski, R.~H. Campbell, K.~Czechowski,
  D.~Erhan, C.~Finn, P.~Kozakowski, S.~Levine, A.~Mohiuddin, R.~Sepassi,
  G.~Tucker, and H.~Michalewski, ``Model based reinforcement learning for
  atari,'' in \emph{8th International Conference on Learning Representations
  (ICLR)}, 2020.

\bibitem{dugas2022navdreams}
D.~Dugas, O.~Andersson, R.~Siegwart, and J.~J. Chung, ``Navdreams: Towards
  camera-only rl navigation among humans,'' in \emph{IEEE/RSJ International
  Conference on Intelligent Robots and Systems (IROS)}, 2022, pp. 2504--2511.

\bibitem{brohan2022rt}
A.~Brohan, N.~Brown, J.~Carbajal, Y.~Chebotar, J.~Dabis, C.~Finn,
  K.~Gopalakrishnan, K.~Hausman, A.~Herzog, J.~Hsu, \emph{et~al.}, ``Rt-1:
  Robotics transformer for real-world control at scale,'' \emph{arXiv preprint
  arXiv:2212.06817}, 2022.

\bibitem{brohan2023rt}
A.~Brohan, N.~Brown, J.~Carbajal, Y.~Chebotar, X.~Chen, K.~Choromanski,
  T.~Ding, D.~Driess, A.~Dubey, C.~Finn, \emph{et~al.}, ``Rt-2:
  Vision-language-action models transfer web knowledge to robotic control,''
  \emph{arXiv preprint arXiv:2307.15818}, 2023.

\bibitem{keipour2022visual}
A.~Keipour, G.~A. Pereira, R.~Bonatti, R.~Garg, P.~Rastogi, G.~Dubey, and
  S.~Scherer, ``Visual servoing approach to autonomous uav landing on a moving
  vehicle,'' \emph{Sensors}, vol.~22, no.~17, p. 6549, 2022.

\bibitem{cho2022autonomous}
G.~Cho, J.~Choi, G.~Bae, and H.~Oh, ``Autonomous ship deck landing of a
  quadrotor uav using feed-forward image-based visual servoing,''
  \emph{Aerospace Science and Technology}, vol. 130, p. 107869, 2022.

\bibitem{wang2023image}
G.~Wang, J.~Qin, Q.~Liu, Q.~Ma, and C.~Zhang, ``Image-based visual servoing of
  quadrotors to arbitrary flight targets,'' \emph{IEEE Robotics and Automation
  Letters}, vol.~8, no.~4, pp. 2022--2029, 2023.

\bibitem{ge2022vision}
R.~Ge, M.~Lee, V.~Radhakrishnan, Y.~Zhou, G.~Li, and G.~Loianno, ``Vision-based
  relative detection and tracking for teams of micro aerial vehicles,'' in
  \emph{IEEE/RSJ International Conference on Intelligent Robots and Systems
  (IROS)}, 2022, pp. 380--387.

\bibitem{szeliski2022computer}
R.~Szeliski, \emph{Computer Vision: Algorithms and Applications}, 2nd~ed., ser.
  Texts in Computer Science.\hskip 1em plus 0.5em minus 0.4em\relax Cham:
  Springer International Publishing, 2022.

\bibitem{loianno2016estimation}
G.~Loianno, C.~Brunner, G.~McGrath, and V.~Kumar, ``Estimation, control, and
  planning for aggressive flight with a small quadrotor with a single camera
  and imu,'' \emph{IEEE Robotics and Automation Letters}, vol.~2, no.~2, pp.
  404--411, 2016.

\bibitem{Jocher_YOLO_by_Ultralytics_2023}
\BIBentryALTinterwordspacing
G.~Jocher, A.~Chaurasia, and J.~Qiu, ``{YOLO by Ultralytics},'' Jan. 2023.
  [Online]. Available: \url{https://github.com/ultralytics/ultralytics}
\BIBentrySTDinterwordspacing

\bibitem{tensorrt}
\BIBentryALTinterwordspacing
{NVIDIA Corporation}, ``Nvidia tensorrt,'' Year of the version you're
  referencing, e.g., 2021. [Online]. Available:
  \url{https://developer.nvidia.com/tensorrt}
\BIBentrySTDinterwordspacing

\bibitem{lin2014microsoft}
T.-Y. Lin, M.~Maire, S.~Belongie, J.~Hays, P.~Perona, D.~Ramanan,
  P.~Doll{\'a}r, and C.~L. Zitnick, ``Microsoft coco: Common objects in
  context,'' in \emph{Computer Vision -- ECCV 2014}, D.~Fleet, T.~Pajdla,
  B.~Schiele, and T.~Tuytelaars, Eds.\hskip 1em plus 0.5em minus 0.4em\relax
  Cham: Springer International Publishing, 2014, pp. 740--755.

\end{thebibliography}

\end{document}